\documentclass{ecai} 
\usepackage{latexsym}
\usepackage{amssymb}
\usepackage{amsmath}
\usepackage{amsthm}
\usepackage{booktabs}
\usepackage{enumitem}
\usepackage{graphicx}
\usepackage{color}
\usepackage{xspace}
\usepackage{multirow}
\usepackage{makecell}
\usepackage{hyperref}
\usepackage{cleveref}
\RequirePackage{algorithm}
\RequirePackage{algorithmic}

\newcommand{\BibTeX}{B\kern-.05em{\sc i\kern-.025em b}\kern-.08em\TeX}

\DeclareMathOperator*{\softmax}{softmax}
\DeclareMathOperator*{\score}{score}
\DeclareMathOperator*{\argmin}{argmin}
\DeclareMathOperator*{\acc}{acc}
\newcommand{\CConf}{\ensuremath{M_D}}
\newcommand{\ConfBack}{\text{Conf$_B$}}
\newcommand{\ConfForward}{\text{Conf$_F$}}
\newcommand{\jchosen}{\ensuremath{j^*}}

\newcommand{\bidirShort}{\textsc{Bi-Dir}\xspace}
\newcommand{\auxShort}{\textsc{AuxFor}\xspace}
\newcommand{\backwardsShort}{\textsc{BackInt}\xspace}
\newcommand{\maxPShort}{\textsc{MaxProb}\xspace}
\newcommand{\entBinShort}{\textsc{EntBin}\xspace}
\newcommand{\binAccShort}{\textsc{BinAcc}\xspace}

\newcommand{\randShort}{\textsc{Rand}\xspace}

\begin{document}

%%%%%%%%%%%%%%%%%%%%%%%%%%%%%%%%%%%%%%%%%%%%%%%%%%%%%%%%%%%%%%%%%%%%%%%%

\begin{frontmatter}

%%% Use this command to specify your submission number.
%%% In doubleblind mode, it will be printed on the first page.

\paperid{123} 

%%% Use this command to specify the title of your paper.

\title{Bi-directional Model Cascading with Proxy Confidence}
%%% Use this combinations of commands to specify all authors of your 
%%% paper. Use \fnms{} and \snm{} to indicate everyone's first names 
%%% and surname. This will help the publisher with indexing the 
%%% proceedings. Please use a reasonable approximation in case your 
%%% name does not neatly split into "first names" and "surname".
%%% Specifying your ORCID digital identifier is optional. 
%%% Use the \thanks{} command to indicate one or more corresponding 
%%% authors and their email address(es). If so desired, you can specify
%%% author contributions using the \footnote{} command.

\author[A]{\fnms{David}~\snm{Warren}\orcid{0000-0002-4914-1344}\thanks{Corresponding Author. Email: david.warren@hdr.mq.edu.au.}}
\author[A]{\fnms{Mark}~\snm{Dras}\orcid{0000-0001-9908-7182}}

\address[A]{Macquarie University, Sydney, Australia}

%%% Use this environment to include an abstract of your paper.

\begin{abstract}
Model Cascading, recently applied successfully to LLMs, is a simple but powerful technique that improves the efficiency of inference by selectively applying models of varying sizes. Models are used in sequence from smallest to largest, only deferring samples to large, costly models when smaller models are not sufficiently confident. Existing approaches to deferral use only limited small model confidence estimates because of the inaccessibility of the large model, although large model confidence is known to be important. We therefore propose a bi-directional approach to deferral that considers the confidence of small and large models in the cascade simultaneously through the use of a proxy for the large model.  This requires a richer representation of model confidence to enable comparative calibration: we use an analysis of hidden states to improve post-invocation confidence of the small model, which in itself improves cascading results over prior approaches.  We then combine this with a tiny proxy model to estimate pre-invocation confidence of the large model. We examine the proposed cascading system over challenging, multiple-choice datasets, finding improvements over standard cascading baselines reflected in reductions in deferrals to more costly models.
\end{abstract}

\end{frontmatter}

%%%%%%%%%%%%%%%%%%%%%%%%%%%%%%%%%%%%%%%%%%%%%%%%%%%%%%%%%%%%%%%%%%%%%%%%

\section{Introduction}
\label{sec:intro}

In the last decade large, transformer-based, pretrained language models (LLMs) have surged in both popularity and capability, radically improving the state of the art in a wide range of tasks, from traditional Natural Language Processing (NLP) tasks to more general problems that were previously impossible, such as complex logical reasoning and the application of expert knowledge. 
% This expanded capability, has placed LLMs in a unique position as a potentially useful tool in many real-world applications, 
% however, there are barriers that prevent widespread use of the profound capabilities of modern LLMs and many of these barriers stem from the rapidly increasing number of parameters in modern LLMs. These larger models have increased computational requirements, larger memory footprint, and longer delays resulting in higher capital and operational costs, environmental impacts, AI accessibility issues, and many more. In response to these problems, a diverse range of techniques have been developed to accelerate or improve the efficiency of LLMs.
However, these abilities have often required increasingly large model sizes where inference is expensive, raising concerns ranging from user budgets \cite{chen2023frugalgptuselargelanguage} to environmental impacts \cite{singh2024surveysustainabilitylargelanguage,ren-etal:2024:SciRep}, including among the NLP community \cite{arase-etal:2021:ACL}.
In response to this, a diverse range of techniques have been developed to accelerate or improve the efficiency of LLMs.

Some approaches aim to reduce the size of individual models like quantization \cite{NEURIPS2022_adf7fa39} and model pruning \cite{sun2024a}. Others have drawn from the field of adaptive computation, developing systems of models that collaborate to improve some aspect of performance. Conceptually relying on the fact that not all tasks are equally difficult, these approaches make the inference process dependent on the individual sample, leveraging smaller models or subsystems to handle easy instances. This concept has been used to provide large improvements in model throughput using draft-and-verify paradigms \cite{chen2023accelerating, leviathan2023fast} as well as to optimize the accuracy-efficiency trade-off \cite{Nie2024OnlineCL}. Early-exit model cascading is a technique that has been used in other machine learning domains for some time \cite{cascades2001, Park2015BiglittleDN} that takes advantage of this principle by engaging a sequence of models of increasing size and capability. Inference is performed on each sample starting with the smallest model and exiting the cascade if the output of any model appears sufficient by some measure of confidence. This approach has recently been applied to NLP and LLMs and has delivered significant efficiency improvements \cite{yue2024large}. However, there are some notable limitations that we address in this paper. 

In cascading models, there is typically a sequence of models of increasing complexity, $M_1 \ldots M_K$, and \citet{jitkrittum2023when} observe that standardly, the decision to defer from a smaller model $M_i$ to larger model $M_{i+1}$ is based only on some measure of confidence from the smaller model $M_i$. This can result in various failure modes, for example in confidence-based deferral erroneously forwarding samples where $M_{i+1}$ performs worse than $M_i$.  \citet{jitkrittum2023when} carry out a theoretical analysis for the two-model case that defines oracles that include probabilities from the larger model, formally characterising those cases where deferrals based on small-model probabilities alone are insufficient or non-optimal.  For practical use, they define post-hoc estimators of their oracle rules that include theoretical confidence of both small and large models.  Importantly, they note that since querying the larger model defeats the entire purpose of cascades, they use only the outputs of the small model in their estimators.  Their experiments on computer vision datasets, however, do not show much improvement over the standard use of small model confidence.

The first core idea of the present paper is to use a proxy for large model confidences that can be obtained without querying the large model, by building a much smaller model whose only role is to predict large model confidences.  For this, we draw on recent work on calibrating LLMs by \citet{ulmer-etal-2024-calibrating} that will allow us to use only model inputs.  We then have `backward', post-invocation confidence for the smaller model, and a `forward', pre-invocation 'proxy confidence' estimation for larger models using a small auxiliary model.

There is then the question of how to relate these two kinds of confidence estimation.  We cannot expect that they will be calibrated with respect to each other: in the first place, one is directly taken from the inference model and the other from a proxy; in addition, it is known that larger models produce better calibrated results than smaller models \cite{zhu-etal-2023-calibration}.  Our second core idea is then to use a meta-model for making deferral decisions in a cascade, extending their use from single-model calibration \cite{pmlr-v89-chen19c,shen-etal:2023:AAAI}.  This then requires a richer representation of features for model confidence: where existing cascading work largely uses final-layer maximum softmax probability output or entropy, we use intermediate layer probing results, which have been shown to be effective in improving calibration in various contexts \cite{papernot2018deep, mielke-etal-2022-reducing,chuang2024dola}. 

We apply our approach to challenging NLP benchmark problems where there is a notable benefit to be gained from deferring to a larger model. 
Our contributions are:

\begin{itemize}
    \item We present a new bi-directional approach to cascading with a meta-model incorporating (i) richer representations of multiple model confidences using internal model states, and (ii) novel proxy confidence measures that avoid queryring large models.

    \item We find the bi-directional approach outperforms methods that use only smaller model confidence.  We also find that, by itself, using internal state representations for smaller model confidence improves over the standard use of final layer maximum probability.
    
    \item The improvements translate to reductions in data items deferred to the larger model of up to 42.5\%.
    
\end{itemize}

%%%%%%%%%%%%%%%%%%%%%%%%%%%%%%%%%%%%%%%%%%%%%%%%%%%%%%%%%%%%%%%%%%%%%%%%

\section{Related Work}
\label{sec:rel-work}

The accuracy-efficiency trade-off is a pervasive concern in ML, particularly in the field of LLMs, and there are many methods for reducing the computational requirements of models while minimally compromising on output quality \cite{kaplan2020scalinglawsneurallanguage}. The idea of making an early-exit from  inference is one that has existed for many years \cite{cascades2001}, but has seen renewed interest with the increasing size and commercial viability of modern models \cite{earlyexitsurvey}. Broadly this involves managing the interaction between each sample and the inference system to minimize wasteful computation. This can been applied at the model-level, directing the flow of samples through a set of discrete models, or at the layer-level, controlling the number of dependent layers to be used within a single neural network model \cite{miao2024efficientinferenceframeworkearlyexit}. 

\subsection{Model Cascades}
In early-exit model cascades inference is first performed on a sample using the smallest available model, then some estimate of the confidence of the result is used to determine whether the sample should be deferred to the next largest model or the output should be accepted as is. This process occurs iteratively until the output is accepted or the largest model in the cascade is invoked \cite{varshney-baral-2022-model}. The decision to accept or defer is typically a comparison between the confidence and some static or dynamic threshold, which can be adjusted to satisfy the desired trade-off \cite{Wang2017IDKCF,Lebovitz2023EfficientIW}. 

Approaches in existing literature vary in number and type of models in the cascade. \citet{Lebovitz2023EfficientIW} find that two model cascades perform well, with three model cascades performing marginally better at the cost of increased system complexity, while others \cite{chen2023frugalgptuselargelanguage} include many models to take advantage of disparate commercial API prices. \citet{Nie2024OnlineCL} cascade 3 heterogeneous models (Logistic Regression, BERT and GPT/LLaMA) with online optimization, treating the largest model outputs as ground-truth with strong results; however, this constrains the application to tasks where traditional models (LR) are viable and assumes the success of largest available model. Choice of confidence metric $A$ also varies, with maximum probability and entropy both being frequently used and effective with some conflicting results regarding which is better \cite{Lebovitz2023EfficientIW, Wang2017IDKCF}.
Our work broadly fits into this part of the space, following the observations of \citet{jitkrittum2023when} about the potential (but not necessarily realised) value of using confidence from multiple models in a cascade.

Other recent work has focussed on LLM-specific features. To note three: \citet{gupta2024language} use language model cascades to accomplish generative tasks; \citet{yue2024large} use Chain-of-Thought prompting to improve small model confidence; \cite{wang2024cascadeawaretraininglanguagemodels} alter the LLM training paradigm to produce models best suited to cascading.

%A task related to model cascading is Learning to Defer (L2D) \citep{madras-etal:2018:NeurIPS}; \citet{jitkrittum2023when} discusses their relationship.  In summary, most work in the L2D space assumes a single stage framework involving training models from scratch as part of simultaneous learning of predictor and deferral function, which is not feasible for LLMs \citet{mao2023twostage,narasimhan-etal:2022:NeurIPS}.  The two-stage approaches do still assume access to large model output in the loss function for learning deferral, in contrast to our scenario. 

A task related to model cascading is Learning to Defer (L2D) \citep{madras-etal:2018:NeurIPS}; \citet{jitkrittum2023when} discusses their relationship.  In summary, most work in the L2D space assumes a single stage framework involving training models from scratch as part of simultaneous learning of predictor and deferral function, which is not feasible for LLMs \citet{mao2023twostage,narasimhan-etal:2022:NeurIPS}.  More generally,
work in the L2D space is aimed at developing a framework in which models can be trained to make a prediction or defer to a (usually human) expert, using a combined cost function including some expert invocation penalty. \citet{narasimhan-etal:2022:NeurIPS} note that L2D and model cascading can be similar under some constraints; however, traditional cascading approaches like ours, model only the expected performance improvement of deferral, allowing the user to determine the cost-performance tradeoff in a post-hoc manner without retraining.

%the l2d considers cost of expert invocation in the loss function for model or deferral model training, whereas we assume only that smaller model is preferrable when correct, and allow the user to determine the relationship between performance and expert invocation cost post-hoc

\subsection{LLM Calibration}
A model's calibration is its ability to express some level of confidence associated with an output that accurately reflects the likelihood that it is correct \cite{pmlr-v70-guo17a}. LLM calibration generally has great value to researchers and industry applications, and is the subject of significant research efforts \cite{geng-etal-2024-survey}. The acceptance of LLMs in real-world application depends on their trustworthiness and reliability \cite{bhatttruthfulness}, while in cascading context, calibration has direct and measurable benefits on performance. The success of any model cascade depends on the quality of its deferral process, and therefore the calibration of its component models or accuracy of its confidence metric $A$. 

There are many approaches to approximating LLM confidence. A large subset of these methods involves deriving the confidence from the log-probabilities of the model outputs. In the case of LLMs, these log-probabilities, or ``logits'', reflect a distribution over tokens and the confidence can be estimated using some variation of entropy or maximum probability at the individual token or sequence level. Some techniques that have shown recent promise involve examining consistency across multiple stochastic or permuted generations \cite{huang}, but the additional computation required makes these approaches poorly suited to efficiency optimizations. Other researchers have asked generative models to explicitly report their confidence in their output with mixed results \cite{kadavath, pawitan2024confidencereasoninglargelanguage}, but benchmarks in some domains have found self reported confidence to be very poor, especially among smaller models \cite{Omar2024.08.11.24311810}.

Different layers in a language model capture different types of information and fulfill different roles in inference \cite{tenney-etal-2019-bert}. One method of identifying and promoting fact-based and truthful outputs from LLMs is to consider the likelihood of different tokens at different layers within the model, emphasizing the effect of layers most likely to inject knowledge \cite{chuang2024dola}. The internal layers of a pretrained model inform output rather than generate output directly and may hold useful information without the same biases as the output layer \cite{Ji2022SurveyOH}. Other, similar techniques applying an output head over internal layers have been used in early-layer exit strategies \cite{teerapittayanon2017branchynetfastinferenceearly}, improve interpretability \cite{explanability}, and to better understand how the internal layers and mechanisms function in LLMs \cite{he-etal-2024-decoding, tighidet-etal-2024-probing}.

In the APRICOT method of \citet{ulmer-etal-2024-calibrating}, a small auxiliary language model is used in support of a larger LLM to approximate the confidence of the larger model based on an embedding of the question and the model's output. They demonstrate strong calibration scores and superior identification of mistakes compared to logit based sequence likelihood and self-assessed confidence. Further, ablations in this study indicate that confidence can reasonably be estimated without model outputs, a potential boon for cascading systems where confidence prior to execution is more valuable for efficiency.

%%%%%%%%%%%%%%%%%%%%%%%%%%%%%%%%%%%%%%%%%%%%%%%%%%%%%%%%%%%%%%%%%%%%%%%%%%%%
\section{Methodology}
\label{sec:method}

\begin{figure}[t]
\begin{center}
\centerline{\includegraphics[width=\columnwidth\relax]{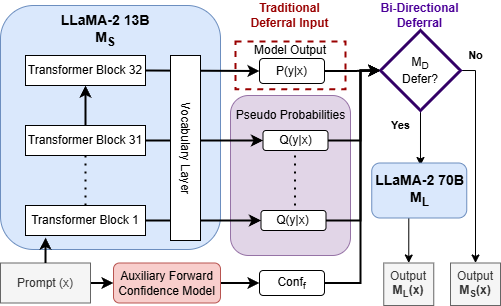}}
\caption{Bi-Directional deferral system diagram.}
\label{backint-diagram}
\end{center}
\end{figure}

\subsection{Cascade Definition}

As noted in Sec~\ref{sec:intro}, existing cascade work generally defines the deferral choice only in terms of one model.  Here we define our cascade in a form similar to that of \citet{varshney-baral-2022-model} but in terms of confidence based on two adjacent models in the cascade.

A cascading system $S$ is more formally, for a sample $x \in \mathcal{X}$ and some set of $K$ models ($M_1, \ldots, M_K$) ordered by the cost of inference on that sample such that $C^x_1 < \ldots < C^x_k$, controlled by a deferral function $D^x$ that determines whether the output from a model should be used as the system output or the sample should be deferred to larger models. Our deferral function is derived from a function $\CConf$ that considers estimates of confidence from both the model in focus $M_j$ (confidence representation $\ConfBack$) and the next model in the cascade $M_{j+1}$ (confidence representation $\ConfForward$) at sample $x$, relative to some threshold $\tau_j$. Models are called from smallest to largest, with $D^x_j = 1$ indicating deferral for model $M_j$ until a model is chosen, as described in Algo~\ref{alg:cascade}.  The selected model is then $M_{\jchosen}$ where $\jchosen = \argmin_j \{ D^x_j = 0 \}$.
The output of the system is then simply the accepted model output:
$S(x) = M_{\jchosen}(x)$.

% $$D^x_j = \begin{cases}
%   1, & \text{if } A(M_j(x)) < \tau, \text{ reject the output of }M_j \text{ defer to } M_{j+1} \\
%   0, & \text{if } A(M_j(x)) >= \tau, \text{ accept output, select }j_x =j \text{ and exit} .
% \end{cases}$$

% $$D^x_j = \begin{cases}
%   1, & \parbox{0.75\columnwidth}{if $\CConf(\ConfBack(M_j(x)),\ConfForward(M_{j+1}(x)) < \tau_j$, reject the output of $M_j$, defer to $M_{j+1}$} \\
%   0, & \parbox{0.75\columnwidth}{if $A(M_j(x)) >= \tau$, accept output, select $j_x =j$ and exit}
% \end{cases}$$

\begin{algorithm}[t]
\begin{algorithmic}[1]
\REQUIRE $K \geq 2$ models $M_1, \ldots, M_K$,
backward confidence representation $\ConfBack(\cdot)$,
forward confidence representation $\ConfForward(\cdot)$,
deferral model $\CConf(\cdot, \cdot)$,
thresholds $\tau_1, \ldots, \tau_{K-1}$
\REQUIRE An input instance $x \in \mathcal{X}$

\ 

\FOR{$j = 1, \ldots, K$}
  \STATE $D^x_j = 0$
\ENDFOR

\FOR {$j = 1, \ldots, K-1$}
  \IF {$\CConf(\ConfBack(M_j(x)), \ConfForward(M_{j+1}(x))) > \tau_j$}
    \STATE $D^x_j = 1$
  \ELSE
    \STATE \textbf{break}
  \ENDIF
\ENDFOR

\end{algorithmic}
\caption{Cascade algorithm}
\label{alg:cascade}
\end{algorithm}

% The average cost of inference $c_X$ for a dataset $X$ comprised of $N$ samples $x_1, ...x_n$  as the sum of costs up to $C^x_{\jchosen}$:

% $$c_X = \frac{\sum_{x\in X}\sum_{i=1}^{j_x}C^x_i}{N}$$

% Similarly, average 
Average system performance $s_{\mathcal{X}}$ for the dataset $\mathcal{X}$ is dependent on the output of selected model for each sample $M_{\jchosen}$ using any given evaluation function $\score(\cdot)$:

$$ s_{\mathcal{X}}  = \frac{\sum_{x\in \mathcal{X}}{\score(M_{j^*}(x))}}{N} $$

While our methods apply to cascades of more than two models, in the following we will only work with two models, a small one $M_S (= M_1)$ and a large one $M_L (= M_2)$.
Figure~\ref{backint-diagram} gives a schematic of our overall cascade.

\subsection{Backwards Confidence Estimation}
\label{sec:model-conf-bkwd}

The post-invocation small model confidence method uses internal representations to understand how an output was influenced by the different layers in the model, and uses them to represent how reliable the output is. The general idea of using internal representations for better calibrated probabilities has been applied in \citet{beigi-etal-2024-internalinspector} and \citet{azaria-mitchell-2023-internal}.
% \MDnote{240117}{Some citations. DW: Citations added}. 
Our formulation below is similar to that of \citet{chuang2024dola}, applied there to improve factuality, and here adapted for classification tasks.  We apply it to develop better representations for confidence.

In modern LLMs an embedding layer transforms a sequence of tokens into a sequence of vectors, which are then passed through a set of $N$ transformer blocks, followed by an affine layer $\phi(\cdot)$ at the end for predicting the next-word distribution. Given an input $x \in \mathcal{X}$, we denote the output of the $j$th block by $ H_j = \{ h_1^{(j)}, \ldots, h_{t-1}^{(j)} \}$.  $H_0$ is the embedding layer, and we refer to other $H_j$ where $j < N$ as intermediate layers or hidden states.

In general, at the final transformer block output $h^{(N)}$, the affine layer $\phi$ is typically scaled using a softmax function giving what can be interpreted as a probability distribution over the model $M$'s label set $\mathcal{Y}$:

\[
{p_M(y|x) = \softmax(\phi(h^{(N)}))}, \ \ y \in \mathcal{Y}
\]

We additionally use internal representations where we apply the affine vocabulary layer to selected hidden states, calculating the softmax of the log-likelihood of the candidate tokens internally.  
% Also, as like most other work on cascades we focus on classification tasks in this paper, our $x_t$ will be the class label and will come from a small range of options; we therefore identify the set of the $K$ most likely candidate output tokens $V_C = \{ x_{C_1}, \ldots, x_{C_K} \} \subseteq V$ from the final output distribution given input sequence $x_1, \ldots, x_{t-1}$. 
This produces a probability-like distribution $q_{M,j}$ for candidate tokens at each layer $j$, given by:

$$ {q_{M,j}(y|x) = \softmax(\phi(h^{(j)}))}, \ \ y\in \mathcal{Y}, j \in (\mathcal{J} \cup \{ N \}), $$

where $\mathcal{J}  \subseteq \{ 0, 1, \ldots, N-1 \}$ is a set of candidate internal layers.
The confidence representation $Q_M(x)$ will then consist of a vector of these pseudo-probabilities $q_{j}(y|x)$ for some $\mathcal{J}$, of length $| \mathcal{Y} | \times (| \mathcal{J} | + 1)$, to be used as the basis for features in the deferral model of Sec~\ref{sec:model-deferral}.
Using multiple internal layers provides an indicator of confidence for use in the deferral process by capturing the depth at which each potential response became more or less likely and the stability of each candidate output through the model. 

We then apply some function $g$ to $Q_M$ to derive the backward features for the small model $\ConfBack(M_S(x)) = g(Q_{M_S}(x))$ (Algo~\ref{alg:cascade}) used in our deferral model (Sec~\ref{sec:model-deferral}).  The specific functions we consider in this paper are:

\begin{itemize}
    \item $g_{\text{ident}}(Q_M(x)) = Q_M(x)$ (i.e., the identity);

    \item $g_{\text{maxP}}(Q_M(x)) = \max(q_{M,N}(y|x))$; and

    \item $g_{\text{ent}}(Q_M(x)) = H(q_{M,N}(y|x))$,
\end{itemize}

\noindent
where $H(\cdot)$ is Shannon entropy.  Existing cascade models are then generally special cases of this, with most just using $g_{\text{maxP}}(\cdot)$ (i.e., the maximum score from the final layer, and comparing to a threshold), noted by \citet{gupta2024language} as being widely used as it is ``remarkably hard to surpass in most natural settings''.

\subsection{Forwards Confidence Estimation}
\label{sec:model-conf-fwd}

Ideally, we would be able to derive the same backwards confidence representation for the large model.  However, as noted by \citet{jitkrittum2023when}, in this kind of cascading scenario we do not have access to the large model during inference to obtain these.  Our idea, then, is to predict values that can take the place of these.

With the goal of calibrating LLMs in a QA context using their generations only, \citet{ulmer-etal-2024-calibrating} presented a method for setting confidence targets and training an additional model that predicts an LLM’s confidence based on its textual input and output alone.  In our scenario, we do not have access to large model output, but do have access to the input; the ablation analysis in Appendix A.5 of \citet{ulmer-etal-2024-calibrating} suggests that input text alone could be feasible as the input to the additional predictor model. 

Our pre-invocation large model confidence method thus uses a small auxiliary support model to learn the relationship between an input sequence and the expected large model output confidence. In order to train this model, we first produce a dataset by running once-off inference on a subset of the task data using the large model $M_L$. 
Following \citet{Zadrozny2001ObtainingCP},
entropy is calculated over the outputs of the model and samples are divided into bins by their entropy rank. Mean accuracy of samples in each bin provides a mapping from entropy to direct calibration target, then mean bin accuracy for each sample is used as the training target for the auxiliary forward confidence estimator. 

 For a set of training samples $\mathcal{X}_F$, we first obtain a prediction $\hat{y_i}$ and entropy $g_{\text{ent}}(\cdot)$ using $M_L(x_i)$, $x_i\in \mathcal{X}_F $. We then divide the range of entropy $H$ into 10 equally sized intervals $I_{1\ldots 10}$, assigning each sample $x_i$ to a bin $m$ if its entropy is within the interval $I_m$. Let $B_m$ then be the samples of $X$ that are assigned to bin $m$, we thus obtain mean accuracy of bins using the true class label $y_i$ by:

%\vspace{-0.35cm}

 $$ \acc(B_m) = \frac{1}{|B_m|}\sum_{i \in B_m}{(\hat{y_i} = y_i)}$$

%\vspace{-0.25cm}

We fit the auxiliary model $M_A: \mathcal{X}_F \rightarrow\mathbb{R}$ using with training target $\acc(B_m)$ for samples $x_i \in B_m$ minimizing the Mean Square Error (MSE) loss: 

%\vspace{-0.35cm}

$$ \min_{M_A: \mathcal{X}_F \rightarrow\mathbb{R}}\frac{1}{|\mathcal{X}_F|}\sum_{x_i \in \mathcal{X}_F}\text{MSE}(\acc(B_m), M_A(x_i)) $$

%\vspace{-0.25cm}

The forward feature for the large model is then $\ConfForward(M_L(x)) = M_A(x)$.

\subsection{Deferral Model}
\label{sec:model-deferral}
The traditional cascading concept treats its constituent models as a set of monotonically improving classifiers, where progression to a larger model is contingent on the confidence of the previously invoked models is below some threshold $\tau$. However, a more practical conceptualization of the deferral task is to model the probability that a given sample is better handled by a larger model to the extent that increased cost is justified. So we consider the deferral task not simply uncertainty of the small model, but an estimate of the probability that both the small model is incorrect and the larger model is correct, or the confidence that the sample is a potential gain through cascading.

Since the representation of meaning throughout the model is itself complex and not yet fully understood \cite{jin2024exploring}, we feed all of the extracted pseudo-probabilities along with the forward confidence estimate to a deferral model $M_D$, which aims to identify samples that are most likely to be potential gains through deferral. Since the inputs to this model are numeric rather than text, traditional classifiers are adequate with minimal computational cost. We first define the `gain' binary target, $y_{G,i}$, as samples where the output of small model $M_S$ does not match the true label and the output of large model $M_L$ does match the true label in the once-off generated training data:

%\vspace{-0.35cm}

%$$y_{G,i} = (M_S(x_i) \ne y_i) \land (M_L(x_i) = y_i) $$
$$ y_{G,i} =\begin{cases}1 & \text{if } (\mathsf{MS}(x_i) \ne y_i) \land (\mathsf{ML}(x_i) = y_i) \\
0 & \text{otherwise} \end{cases} $$
%\vspace{-0.25cm}

We then train the deferral model using as features from the small model in the form $\ConfBack$ (Sec~\ref{sec:model-conf-bkwd}), and features from the large model in the form $\ConfForward$ (Sec~\ref{sec:model-conf-fwd}).

Since the deferral model produces probabilistic outputs, we can trivially apply a threshold $0<\tau<1$ at which potential gains should be deferred to a larger model $M_L$ if $M_D > \tau$ (Algo~\ref{alg:cascade}).

\section{Experimental Setup}
\label{sec:exper-setup}

\subsection{Models}
The publicly available pre-trained LLaMA-2 models are used as the cascade component LLMs \cite{Touvron2023Llama2O}, with the 13 billion parameter version used as the smallest cascade model, $M_S$, and the 70 billion parameter LLaMA-2 model as the large model, $M_L$. These models are open and accessible enough to conduct relevant experiments while being modern in architecture and competitive in performance. The primary models do not undergo any training or fine-tuning and are used as-is.

The forward confidence estimator is a pre-trained DeBERTa V3 model with a regression head at the base size with 86 million parameters \cite{he2023debertav3improvingdebertausing}. This model is approximately 0.7\% the size of the smaller cascade model, resulting in a tiny increase in computation for the system. 

The deferral model is a simple Random Forest Classifier using the Scikit Learn library \cite{scikit-learn}, trained on pseudo-probabilities from the backwards confidence estimator and the output of the forward confidence estimator to identify samples where the larger model will provide a better answer than the current output.

All model training uses a 5-fold cross-validation approach and only unseen test fold results are reported.

\subsection{Datasets}
The methods are applied to four commonly used multiple choice question-answering (MCQA) language datasets. We use BoolQ \cite{clark-etal-2019-boolq} yes/no question answering with context as our easiest task, and Massive Multitask Language Understanding (MMLU) \cite{hendrycks2021measuring}, AI2 Reasoning Challenge (ARC) Science exam question dataset (Easy and Challenge subsets; ARC(E) and ARC(C) resp.) \cite{Clark2018ThinkYH}, and Common Sense Question Answering (CSQA) \cite{talmor-etal-2019-commonsenseqa} as more difficult benchmarks.

\subsection{Cascade}
\label{sec:exper-cascade-models}

Our primary method of cascade deferral is using the outputs of the model described in Sec~\ref{sec:model-deferral}, using the full features defined by $g_{\text{ident}}$ in the backwards confidence for the small model combined with $\text{Conf}_F$. We refer to this as the bi-directional model \bidirShort.

As our baselines, we use the following: 

\begin{itemize}

    \item \backwardsShort: This uses only the small model $M_S$, via backward confidence estimation with all internal blocks ($g_{\text{ident}}$).

    \item \entBinShort: This uses only the small model $M_S$, via entropy of the final layer ($g_{\text{ent}}$). This baseline is similar to the DTU method of \citet{varshney-baral-2022-model}.
    
    \item \maxPShort: This uses only the small model $M_S$, via the maximum probability of the final layer ($g_{\text{maxP}}$). This baseline is the MaxProb method of \citet{varshney-baral-2022-model}.

    \item \randShort: This is the baseline of \citet{varshney-baral-2022-model} where the output of $M_S$ or $M_L$ is selected at random.
\end{itemize}

We also include an ``oracle'' version of \bidirShort, where rather than using the proxy forward confidence for the large model $M_L$, we use the actual maximum probability (i.e., $g_{\text{maxP}}(Q_{M_L(x)})$.  (This is a kind of oracle as under the scenario of interest, the cascade would not have access to this in a real-world setting as it requires large model outputs at inference time.)

\begin{table*}[t]
\caption{Deferral AUC scores for methods and baselines across datasets.}
\label{deferral-auc-full-table}
\begin{center}
\begin{small}
\begin{sc}
\begin{tabular}{lrrrrr}
\toprule
 & BoolQ & ARC(E) & ARC(C) & MMLU & CSQA \\
Method &  &  &  &  &  \\
\midrule
\randShort & 0.763621 & 0.825090 & 0.677448 & 0.528871 & 0.464629 \\
\maxPShort & 0.775638 & 0.870865 & 0.707827 & 0.540644 & 0.502895 \\
\entBinShort & 0.773487 & 0.866861 & 0.705555 & 0.540084 & 0.500234 \\
\backwardsShort & \textbf{0.788683} & \underline{0.874897} & \underline{0.715774} & \underline{0.544065} &  \textbf{0.518998} \\
\bidirShort & \underline{0.784171} & \textbf{0.876857} & \textbf{0.722361} & \textbf{0.553650} & \underline{0.517227} \\
\midrule
\bidirShort (Oracle Aux.) & 0.783438 & 0.876916 & 0.722466 & 0.558934 & 0.528311 \\
\bottomrule
\end{tabular}
\end{sc}
\end{small}
\end{center}
\end{table*}

\begin{table*}[t]
\caption{Deferral AUC scores at .2 deferral rate for methods and baselines across datasets}
\label{deferral-auc-2-table}
\begin{center}
\begin{small}
\begin{sc}
\begin{tabular}{lrrrrr}
\toprule
 & BoolQ & ARC(E) & ARC(C) & MMLU & CSQA \\
Method &  &  &  &  &  \\
\midrule
\randShort & 0.151320 & 0.155458 & 0.119951 & 0.094581 & 0.075441 \\
\maxPShort & 0.153015 & 0.160308 & 0.121535 & 0.095532 & 0.079433 \\
\entBinShort & 0.150915 & 0.159416 & 0.121232 & 0.095531 & 0.079020 \\
\backwardsShort & \textbf{0.155369} & \underline{0.162092} & \underline{0.123491} & \underline{0.095540} & \underline{0.081445} \\
\bidirShort & \underline{0.154312} & \textbf{0.162609} & \textbf{0.127639} & \textbf{0.097874} & \textbf{0.081449} \\
\midrule
\bidirShort (Oracle Aux.) & 0.154637 & 0.162779 & 0.127941 & 0.099757 & 0.084447 \\
\bottomrule
\end{tabular}
\end{sc}
\end{small}
\end{center}
\end{table*}

\begin{table}[t]
\caption{Required deferral rate $r_d$ of Bi-Directional method to surpass system accuracy of MaxProb at .2 and .4 deferral rate with estimated total cost reduction in parentheses.}
\label{cost-reduction-table}
\begin{center}
\begin{small}
\begin{sc}
\begin{tabular}{lrrrrr}
\toprule
 & BoolQ & ARC(E) & ARC(C) & MMLU & CSQA \\
$r_d$ &  &  &  &  &  \\
\midrule
\multirow{2}{*}{0.2} & 0.1407 & 0.1444 & 0.0864 & 0.1248 & 0.1775 \\
% & (15.36\%) & (14.40\%) & (29.44\%) & (19.50\%) & (5.83\%) \\
& (15.4\%) & (14.4\%) & (29.4\%) & (19.5\%) & (5.8\%) \\
\multirow{2}{*}{0.4} & 0.1511 & 0.3067 & 0.3130 & 0.2958 & 0.3407 \\
% & (42.49\%) & (15.93\%) & (14.85\%) & (17.78\%) & (10.12\%) \\
& (42.5\%) & (15.9\%) & (14.9\%) & (17.8\%) & (10.1\%) \\
\bottomrule
\end{tabular}
\end{sc}
\end{small}
\end{center}
\end{table}

% \MDnote{250122}{This has implications for which baselines we'll have.  The \citet{varshney-baral-2022-model} baselines don't need this training process, because they have no knowledge of the large model.  \citet{jitkrittum2023when} by contrast use something like this devset (see Table 1 in their paper) by with features only from the small model.  So there'd be an additional baseline from \citet{jin2024exploring} if I've understood the training correctly.}

\subsection{Evaluation}
Our primary evaluation is measuring the effect on system accuracy as a function of rate of deferral, like \citet{jitkrittum2023when}.  We plot deferral curves for individual method analysis, and provide AUC scores derived from these curves to allow comparison across methods.  Note that these AUC scores differ from those obtained from e.g. ROC curves: their absolute values are determined by the accuracies of $M_S$ and $M_L$, and are mostly of use in relative ranking.  
As we are particularly interested in lower deferral rates (i.e., where most of the work is done by the small model and deferral is avoided where possible), we also calculate AUC for sub-intervals of deferral rates $r_d \in [0,0.2]$ and $r_d \in [0,0.4]$.
To aid in interpreting and quantifying improvements indicated by AUC, we also provide the rates of deferral for different models at the same level of system accuracy.

In our secondary evaluation, we evaluate how well calibrated the confidence estimation methods are in order to understand the reasons for the system performance of the various methods.
In addition to reliability diagrams, like \citet{ulmer-etal-2024-calibrating}, we use
Expected Calibration Error (ECE), Smoothed Expected Calibration Error (smECE), Brier score and AUROC as metrics.

% We test the performance of the two novel confidence estimation methods and compare to baseline confidence estimation in terms of calibration.

% We compare to two baselines calibration methods. Maximum Probability - defined as the simple maximum softmax probability over tokens directly from the output of the model. For MCQA tasks, this probability is calculation is constrained to choices appropriate to the task. Entropy Binning - using the entropy over model token outputs we map entropy onto calibrated probabilities using the binning method described by \citet{Zadrozny2001ObtainingCP} and used in auxiliary model training above. 

% Computational cost for inference is measured in floating-point operations (FLOPs) calculated using the equations of \citet{Kaplan2020ScalingLF}. For a model of $P$ total parameters, $n_{layers}$ layers, with $d$ residual stream values to process a sample of $T$ context tokens we use:

% $$\text{FLOPs per Token} = 2P + 2n_{layers}dT$$

% Real per-token FLOP cost for models used appear in the appendix. 

% \subsection{Ablations}

% In addition to the evaluation of the system performance, we perform a number of ablations to identify contributing factors and relative magnitudes of improvement:
% 1. Deferral using only baseline confidence estimations (entropy and max probability).
% 2. Deferral using only backward confidence estimation with pseudo-probabilities.
% 3. Deferral using baseline (entropy based) backwards estimation and forward confidence estimation.

\section{Results}
\label{sec:results}

\subsection{Main Results: System Performance}

% \MDnote{250130}{(1) In tables, bold highest scores (not including oracle) and underline second highest. (2) Change Method names in tables to macros, e.g. \randShort, \maxPShort, \bidirShort. (3) Change ARC Easy to ARC(E), ARC Challenge to ARC(C).}

Table~\ref{deferral-auc-full-table} presents the AUC scores for the full deferral curves. Our methods provide the strongest deferral performance across all datasets, consistently outperforming naive and informed baselines. \bidirShort achieves best results on on ARC-Easy, ARC-Challenge and MMLU datasets with \backwardsShort performing marginally better on BoolQ and CSQA. We also observe that \backwardsShort is a clear improvement over \maxPShort and other baselines.
Performance of the \bidirShort is very close to the version with oracle forward confidence for BoolQ, ARC-Easy and ARC-Challenge, indicating that the relatively straightforward proxy model is sufficient there.  On the other hand, a more accurate estimate of the large model should improve deferral on MMLU and CSQA, where  $M_L$ is not very accurate; further refinement of forward confidence modeling should lead to improved deferral on these tasks. (For further context, we include a discussion of performance with respect to a full oracle with perfect knowledge in supplementary material ~\ref{app:oracle-deferral}\cite{warren2025bidirectionalmodelcascadingproxy}.)

The Bi-Directional approaches, even with oracle forward confidence, do not outperform the \backwardsShort  method for the BoolQ task. In a boolean context the baseline accuracy of random guessing is 50\%, meaning that calibration should be constrained to a narrower range and may be less informative in both directions. Further, the performance impact of randomly produced correct answers is much greater in this task compared to the other datasets which have chance correctness between 20\% and 25\%.

\begin{figure}[ht]
\begin{center}
\centerline{\includegraphics[width=\columnwidth\relax]{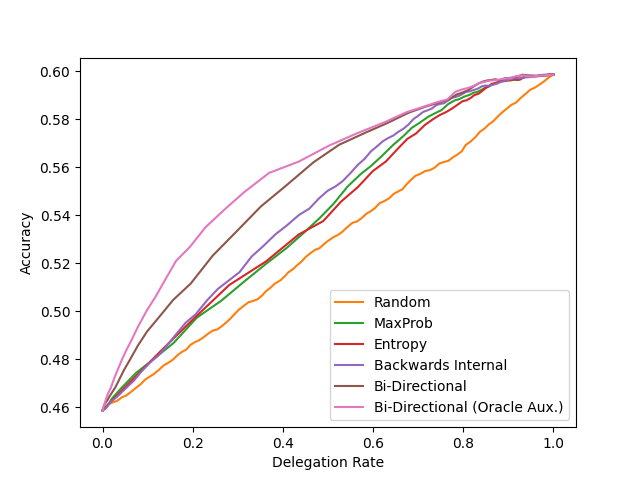}}
\caption{Deferral curve for MMLU showing performance of each method over all delegation rates.}
\label{mmlu-13b-def-curve}
\end{center}
\end{figure}

\begin{figure}[ht]
\begin{center}
\centerline{\includegraphics[width=\columnwidth\relax]{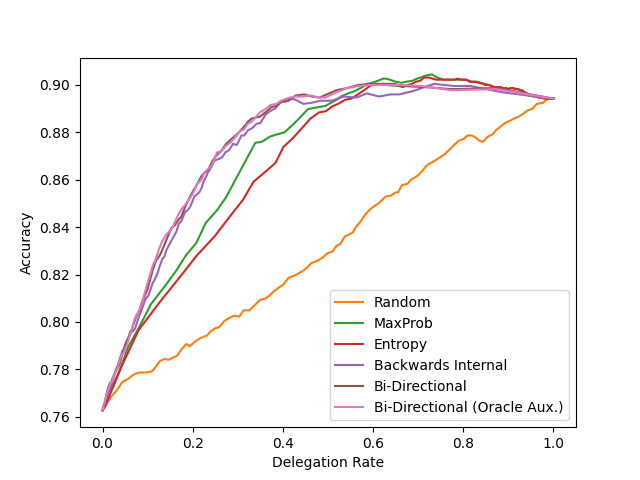}}
\caption{Deferral curve for ARC-Easy showing performance of each method over delegation rates.}
\label{arc-e-13b-def-curve}
\end{center}
\end{figure}

As an illustration of the deferral curves from which the AUC scores are derived, Figure~\ref{mmlu-13b-def-curve} plots cascade system performance against deferral rate using all deferral methods for MMLU. All methods and baselines provide better accuracy gains than chance over the range of delegation rates. The \bidirShort  method outperforms the other methods most in the earlier sections of the curve, avoiding the wasteful deferral of samples likely to be incorrect in both models. This profile is desirable in a typical cascading scenario where the goal is to maximize gains while utilizing the large model sparingly.
Figure~\ref{arc-e-13b-def-curve} shows deferral performance on ARC-Easy, in this case the small model is more capable and the large model correctly classifies almost every sample. This scenario makes \backwardsShort based deferral more competitive with \bidirShort.  While \backwardsShort is effective for tasks that are relatively easy for component models, the \bidirShort method provides efficient deferral for both easy and challenging tasks. Deferral curves for all datasets are included in supplementary material~\ref{app:deferral-curves}\cite{warren2025bidirectionalmodelcascadingproxy}.

Table~\ref{deferral-auc-2-table} presents the AUC scores for the first 20\% of the deferral curves, quantitatively capturing the observation on Figure~\ref{mmlu-13b-def-curve} about important early deferral rates. \bidirShort is the best method on all non-binary tasks with \backwardsShort also consistently outperforming standard baselines. The large gaps between individual small model and large model performance in some datasets means large model performance parity is only obtained at high deferral rates; however, comparison to baselines and oracle deferral show good deferral decisions particularly at low rates for \backwardsShort and \bidirShort. Tasks where small model performance is high---BoolQ, ARC(E)---experience the largest gains above \randShort in both our methods and informed baselines.
AUC scores for the first 40\% of the deferral curves show the same pattern (Table~\ref{deferral-auc-4-table} in supplementary material~\ref{app:deferral-auc}\cite{warren2025bidirectionalmodelcascadingproxy}).

To demonstrate consistency we conducted a non-parametric statistical test of significance using bootstrap resampling and report improvements over the strong MaxProb baseline. Using 1000 test-set resamples with replacement, 95\% CIs indicate significant AUC improvement over baseline (intervals do not include zero) at all operating points for all datasets using the \bidirShort method. The \backwardsShort method alone provides statistically significant improvement at almost every operating point with full results in supplementary material~\ref{app:bootstrap-confidence}\cite{warren2025bidirectionalmodelcascadingproxy}.

To provide an interpretation of these AUC scores, Table~\ref{cost-reduction-table} gives the reduction in items deferred from the small model to the large under \bidirShort relative to \maxPShort while maintaining the same system performance. This reduction in deferral results in a reduction in the total computational cost (small and large model) compared to the strong \maxPShort baseline. Modest changes in deferral AUC in Table~\ref{deferral-auc-full-table} correspond to significant inference cost reductions over \maxPShort delegation. We find lower cost reductions in tasks where both large and small models struggle, particularly CSQA with 5.8\% and 10.1\% reductions for .2 and .4 deferral rates respectively. All other tasks and deferral rates see cost reductions larger than 14\%, with BoolQ at .4 deferral rate exhibiting the highest cost saving of 42.5\%

\subsection{Model Calibration}
\label{sec:results-calibration}

\begin{figure}[h]
\begin{center}
\centerline{\includegraphics[width=\columnwidth\relax]{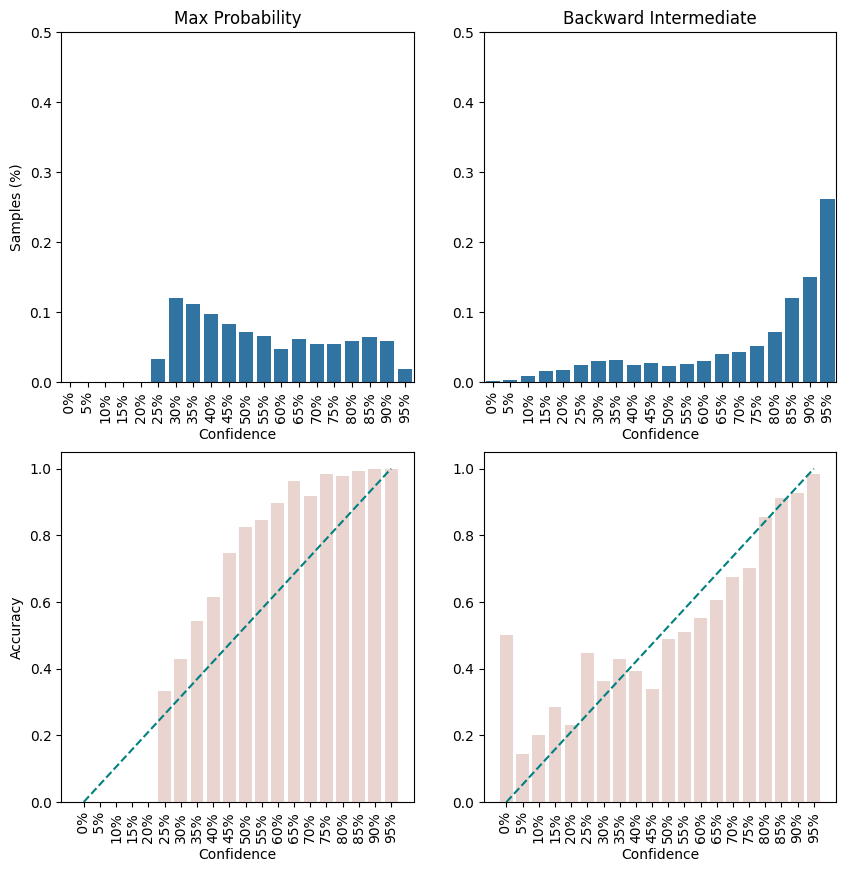}}
\caption{Confidence Histograms and Reliability Diagrams for baseline max probability confidence and internal representations method on the ARC-Easy dataset.}
\label{arc-e-13b-calib}
\end{center}
\end{figure}

In this section we explore why \backwardsShort and \bidirShort produce better results overall by looking at model calibration.
Figure~\ref{arc-e-13b-calib} presents reliability diagrams \citep{pmlr-v70-guo17a} for \backwardsShort calibration alongside \maxPShort for baseline comparison. A key advantage of the \backwardsShort method is that it provides a greater spread of samples across confidence levels, facilitating effective deferral decisions rather than conservatively estimating confidence near the mean accuracy. Specifically in Figure~\ref{arc-e-13b-calib}, \backwardsShort considers a much higher proportion of samples to be high confidence compared to the \maxPShort, and those samples are found to be mostly accurate. This pattern is observed in all datasets to varying degrees (reliability diagrams for remaining datasets appear in supplementary material~\ref{app:backwards-reliability}\cite{warren2025bidirectionalmodelcascadingproxy}).

For an aggregate view, 
Table~\ref{calib-13b-table} in supplementary~\ref{app:backwards-reliability}\cite{warren2025bidirectionalmodelcascadingproxy} presents calibration metrics for the LLaMA-2 13 billion model. Our \backwardsShort method is the best calibrated for all datasets and metrics except for ECE, which is defined similarly to \entBinShort calibration scores. These metrics quantify the degree to which confidence is aligned with correctness but do not measure the ability to represent a wider range of calibrated confidence values. 

Table~\ref{calib-70b-table} in supplementary~\ref{app:backwards-reliability}\cite{warren2025bidirectionalmodelcascadingproxy} shows calibration metrics for the LLaMA-2 70 billion model comparing observed targets to the auxiliary model forward confidence estimation. Auxiliary model confidence is poorly calibrated in general, providing only a rough indicator of expected correctness when compared to observed targets. This rough estimate, however, is available prior to use of the large model and is therefore useful in the deferral process, as evidenced in Table~\ref{deferral-auc-full-table}. We note here that the large and diverse MMLU dataset, which includes questions of varying difficulty and from a wide range of topics can be much more reliably predicted, reaching a correctness AUC of .724 without outputs. 
Auxiliary model performance as a calibrator is weakest on the BoolQ dataset; we discuss this in supplementary material~\ref{app:backwards-reliability}\cite{warren2025bidirectionalmodelcascadingproxy}.

\subsection{Ablations and Further Analyses}

\paragraph{Alternative Auxiliary Target}
In addition to using mean bin accuracy as a target for the auxiliary forward model, we investigate using an auxiliary regression model to directly predict individual output probability from the large model and find the difference to be trivial for both calibration and deferral. The binning method yields slightly better calibration metrics and is the chosen method for all the main experiments.  We provide metrics for both calibration and system performance in Table~\ref{table:aux-model-indiv} in supplementary material~\ref{app:aux-indiv}\cite{warren2025bidirectionalmodelcascadingproxy}.

\paragraph{Auxiliary Model Data Sensitivity}
We examine the sensitivity of the auxiliary forward model to training set size. We conduct the same experiments as above using smaller subsets of the dataset to train the auxiliary model and compare the final deferral performance. While  deferral with access to the larger model output probabilities (Oracle Aux.) performs significantly better than the predicted probabilities at all dataset sizes, we find only small degradations in deferral performance using smaller training set sizes for MMLU down to 25\% of samples. On this diverse and challenging dataset, a small amount of data can provide a useful rough approximation to forward confidence. Table~\ref{data-scaling-def-table} shows forward model data sensitivity for \bidirShort on MMLU where forward confidence was most impactful.

\begin{table}[ht]
\caption{Bi-directional deferral performance using smaller training sets on MMLU.}
\label{data-scaling-def-table}
\begin{center}
\begin{small}
\begin{sc}
\begin{tabular}{lrrr}
\toprule
 & AUC (.2) & AUC (.4) & AUC (full) \\
\midrule
Random & 0.094603 & 0.194561 & 0.528991 \\
\bidirShort(Aux 25\%) & 0.097455 & 0.203643 & 0.552574 \\
\bidirShort(Aux 50\%) & 0.097741 & 0.204144 & 0.552880 \\
\bidirShort(full) & 0.097874 & 0.204550 & 0.553650 \\
\bidirShort(Oracle Aux) & 0.099757 & 0.209051 & 0.558934 \\
\bottomrule
\end{tabular}
\end{sc}
\end{small}
\end{center}
\end{table}

\paragraph{Alternative Small Model}
We also explored the effect of the 7 billion parameter LLaMA-2 model as small model in the cascade. Backwards internal confidence showed improved calibration over standard probability outputs; however, these confidence outputs were generally modest due to the difficulty of the task for the model. The usefulness of a confidence output for the sake of individual sample deferral is limited by its deviation from chance, where high backwards-confidence samples can be retained and low confidence samples deferred. App~\ref{app:calibration-llama-7b} gives details.

\paragraph{Length of Deferred Items}
The size of deferred items will also have an impact on cost: longer items deferred to the larger model will be more costly.
Supplement~\ref{app:prompt-length}\cite{warren2025bidirectionalmodelcascadingproxy} gives an analysis of length of deferred items: \bidirShort delegates shorter items on average. This is likely to be because the deferral model has some knowledge of which items the larger model is uncertain on, which are more likely to be longer.

\section{Conclusion}
\label{sec:conclusions}

We have shown that a model cascade that bases its deferral decisions on richer notions of confidence derived from internal states of smaller  models and from proxy predictions of confidence from larger models can outperform those based on the widely used maximum probability-based confidence. Our method provides consistent improvement in deferral quality over strong baselines in both easy and difficult benchmark tasks. These improvements correspond to reductions in number of items deferred to the large model for the same overall performance, hence reducing inference costs by notable margins, up to 42.5\% on the datasets explored.

Our current work only looked at models from models from the same family; given observations from \citet{Lebovitz2023EfficientIW} regarding the usefulness of cascade diversity, future work could explore relating confidence of different model types. Other future work could involve extending the classification tasks here to generation, which have their own specific confidence-related complexities \citep{gupta2024language}.

%%%%%%%%%%%%%%%%%%%%%%%%%%%%%%%%%%%%%%%%%%%%%%%%%%%%%%%%%%%%%%%%%%%%%%%%

%%% Use this environment to include acknowledgements (optional).
%%% This will be omitted in doubleblind mode.

%\begin{ack}

%\end{ack}

%%%%%%%%%%%%%%%%%%%%%%%%%%%%%%%%%%%%%%%%%%%%%%%%%%%%%%%%%%%%%%%%%%%%%%%%

%%% Use this command to include your bibliography file.
\bibliography{main}

%%%%%%%%%%%%%%%%%%%%%%%%%%%%%%%%%%%%%%%%%%%%%%%%%%%%%%%%%%%%%%%%%%%%%%%%%%%%%%%
%%%%%%%%%%%%%%%%%%%%%%%%%%%%%%%%%%%%%%%%%%%%%%%%%%%%%%%%%%%%%%%%%%%%%%%%%%%%%%%
% APPENDIX
%%%%%%%%%%%%%%%%%%%%%%%%%%%%%%%%%%%%%%%%%%%%%%%%%%%%%%%%%%%%%%%%%%%%%%%%%%%%%%%
%%%%%%%%%%%%%%%%%%%%%%%%%%%%%%%%%%%%%%%%%%%%%%%%%%%%%%%%%%%%%%%%%%%%%%%%%%%%%%%
\newpage

\section{Supplementary Materials}
\appendix

% \onecolumn
\section{Deferral AUC Scores}
% \label{sec:appendix}
\label{app:deferral-auc}

Table~\ref{deferral-auc-4-table} presents the AUC scores for deferral rates between 0 and 0.4.  It is analogous to Table~\ref{deferral-auc-2-table} in the main body, for deferral rates between 0 and 0.2, and shows the same patterns.

\begin{table*}[t]
\caption{Deferral AUC scores at .4 deferral rate for methods and baselines across datasets}
\label{deferral-auc-4-table}
\begin{center}
\begin{small}
% \begin{footnotesize}
\begin{sc}
\begin{tabular}{lrrrrr}
\toprule
 & BoolQ & ARC(E) & ARC(C) & MMLU & CSQA \\
Method &  &  &  &  &  \\
\midrule
\randShort & 0.303772 & 0.315834 & 0.248295 & 0.194945 & 0.158770 \\
\maxPShort & 0.308752 & 0.332378 & 0.254870 & 0.197592 & 0.174113 \\
\entBinShort & 0.306581 & 0.329210 & 0.254853 & 0.198011 & 0.172155 \\
\backwardsShort & \textbf{0.314716} & \underline{0.337152} & \underline{0.260794} & \underline{0.198870} & \underline{0.177947} \\
\bidirShort & \underline{0.312651} & \textbf{0.338187} & \textbf{0.268161} & \textbf{0.204550} & \textbf{0.179089} \\
\midrule
\bidirShort (Oracle Aux.) & 0.312929 & 0.338355 & 0.268365 & 0.209051 & 0.186293 \\
\bottomrule
\end{tabular}
\end{sc}
\end{small}
% \end{footnotesize}
\end{center}
\end{table*}

\section{Deferral Curves}
\label{app:deferral-curves}

\begin{figure}[ht]
\begin{center}
\centerline{\includegraphics[width=\columnwidth\relax]{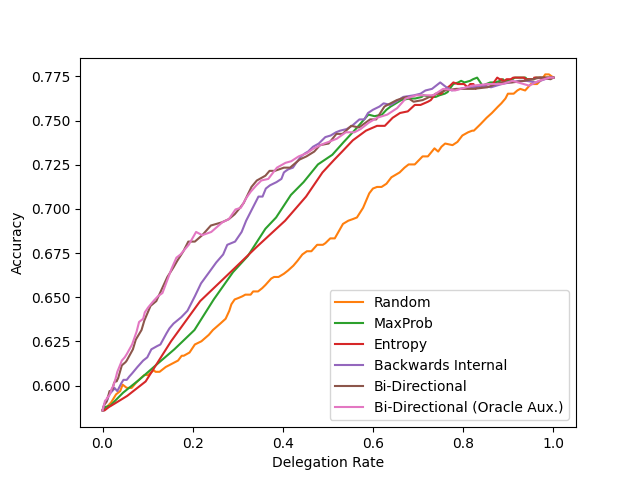}}
\caption{Deferral curve for ARC-Challenge showing performance of each method over all delegation rates.}
\label{arc-c-13b-def-curve}
\end{center}
\end{figure}

\begin{figure}[ht]
\begin{center}
\centerline{\includegraphics[width=\columnwidth\relax]{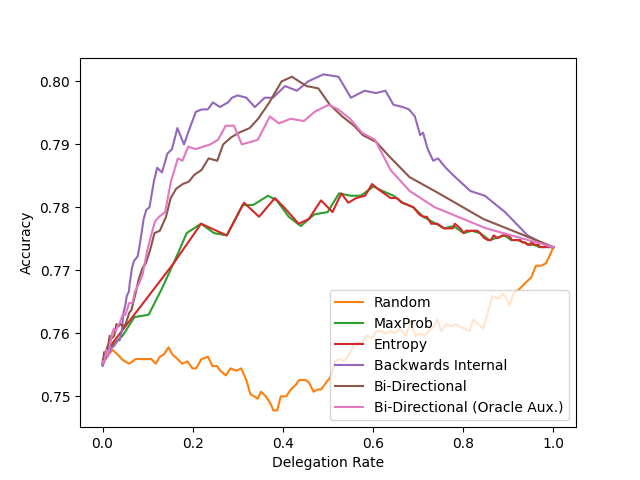}}
\caption{Deferral curve for BoolQ showing system performance over delegation rates.}
\label{boolq-13b-def-curve}
\end{center}
\end{figure}

\begin{figure}[ht]
\begin{center}
\centerline{\includegraphics[width=\columnwidth\relax]{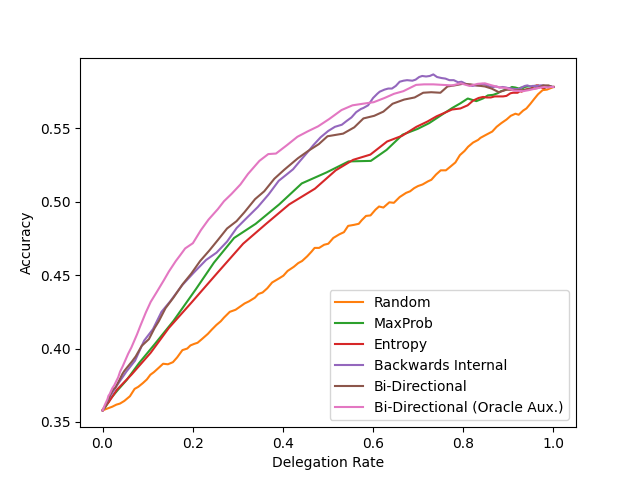}}
\caption{Deferral curve for CSQA showing system performance over delegation rates.}
\label{csqa-def-curve}
\end{center}
\end{figure}

The deferral curves in Figures~\ref{arc-c-13b-def-curve}, \ref{boolq-13b-def-curve}, and \ref{csqa-def-curve} show system performance at all deferral rates for all methods. \backwardsShort and \bidirShort provide the best accuracy-deferral trade-off in all datasets, particularly through the early sections of each curve.

\section{Calibration}
\label{app:backwards-reliability}

\begin{figure}[ht]
\begin{center}
\centerline{\includegraphics[width=\columnwidth\relax]{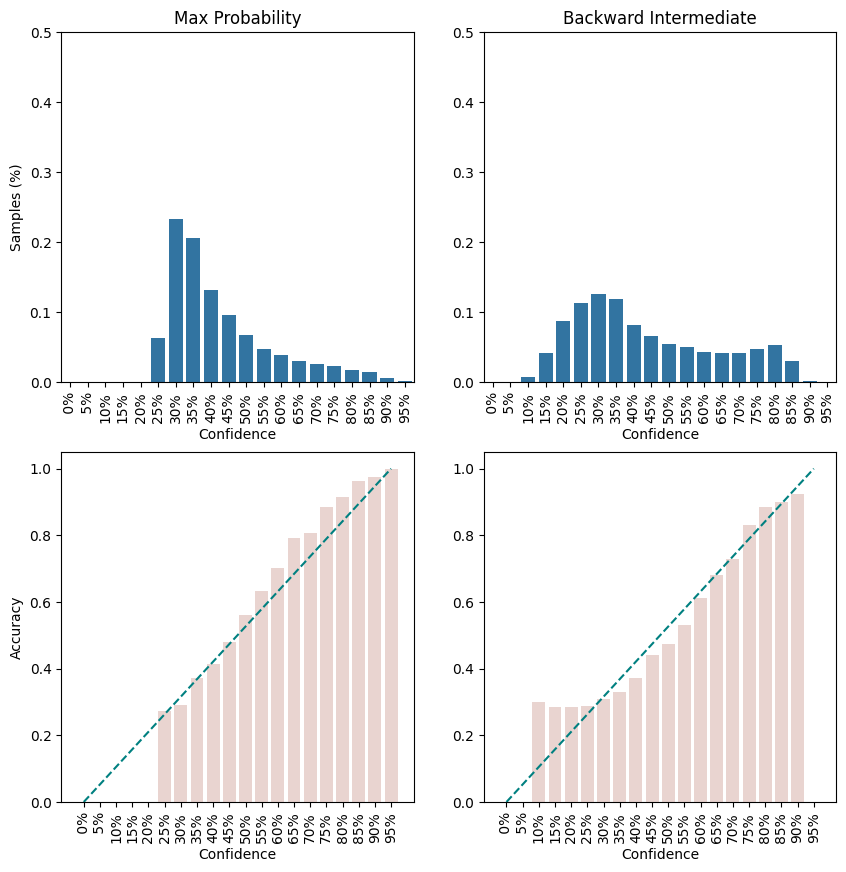}}
\caption{Reliability  diagrams for backwards internal confidence in LLaMA-2 13b on MMLU.}
\label{calib-mmlu}
\end{center}
\end{figure}

Backwards internal representations provide strong calibration metrics as well as more useful reliability diagrams when compared to \maxPShort. The  diagrams demonstrate the tendency of \maxPShort to predict a confidence around the center of the relevant range. While this is not a problem in terms of the ECE, it leads to less informative outputs for many samples and under-confidence in some cases. While \backwardsShort does incorrectly estimate some confidence values under pure chance for the dataset, it provides a much stronger separation of samples into the higher confidence brackets even when the task is relative difficult as in ~\ref{calib-mmlu}.

Auxiliary model performance as a calibrator is generally weak when compared to the observed, post-invocation values but still useful in deferral. It is weakest on the BoolQ dataset, aligning with the weak bi-directional deferral identified in the system level analysis. This weakness may be a result of the probability implications of a binary task, where chance is a large factor, or due to the longer prompts involved in this task. The BoolQ dataset involves relatively large amounts of context information that is useful in answering the binary question, this large context information may prevent the very small auxiliary language model from consistently identifying sample difficulty.

\begin{table}[t]
\caption{Evaluation of Calibration Methods for LLaMA-2 13b model across all datasets.}
\label{calib-13b-table}
\begin{center}
\begin{small}
\begin{sc}
\begin{tabular}{lcccc}
\midrule
 &  \multicolumn{3}{c}{BoolQ} \\
 \midrule
  Method & AUROC$\uparrow$ & Brier $\downarrow$ & ECE$\downarrow$ & smECE$\downarrow$ \\
 \midrule
\makecell{\maxPShort} & 0.714528 & 0.171149 & 0.066581 & 0.184979 \\
\makecell{\entBinShort}& 0.687940 & 0.168369 & \textbf{0.000000} & 0.168958 \\
\makecell{\backwardsShort} & \textbf{0.783132} & \textbf{0.152790} & 0.029252 & \textbf{0.154454} \\
\midrule
 &  \multicolumn{3}{c}{ARC(E)} \\
 \midrule
 Method & AUROC$\uparrow$ & Brier $\downarrow$ & ECE$\downarrow$ & smECE$\downarrow$ \\
 \midrule
\makecell{\maxPShort} & 0.831970 & 0.176012 & 0.186629 & 0.188781 \\
\makecell{\entBinShort} & 0.791273 & 0.144803 & \textbf{0.000000} & 0.145958 \\
\makecell{\backwardsShort} & \textbf{0.854615} & \textbf{0.126268} & 0.036697 & \textbf{0.125896} \\
\midrule
 &  \multicolumn{3}{c}{ARC(C)} \\
 \midrule
 Method & AUROC$\uparrow$ & Brier $\downarrow$ & ECE$\downarrow$ & smECE$\downarrow$ \\
 \midrule
\makecell{\maxPShort} & 0.743045 & 0.215210 & 0.101103 & 0.213175 \\
\makecell{\entBinShort} & 0.720417 & 0.205483 & \textbf{0.000000} & 0.203541 \\
\makecell{\backwardsShort} & \textbf{0.766596} & \textbf{0.192272} & 0.024936 & \textbf{0.188863} \\
\midrule
&  \multicolumn{3}{c}{MMLU} \\
 \midrule
 Method & AUROC$\uparrow$ & Brier $\downarrow$ & ECE$\downarrow$ & smECE$\downarrow$ \\
 \midrule
\makecell{\maxPShort} & 0.698764 & 0.215363 & 0.029530 & 0.216782 \\ 
\makecell{\entBinShort} & 0.671911 & 0.216699 & \textbf{0.000000} & 0.213456 \\
\makecell{\backwardsShort} & \textbf{0.715879} & \textbf{0.208746} & 0.037770 & \textbf{0.206244} \\
\midrule
&  \multicolumn{3}{c}{CSQA} \\
 \midrule
Method & AUROC$\uparrow$ & Brier $\downarrow$ & ECE$\downarrow$ & smECE$\downarrow$ \\
\midrule
\makecell{\maxPShort} & 0.679805 & 0.209745 & 0.063486 & 0.215699 \\
\makecell{\entBinShort} & 0.672926 & 0.205427 & \textbf{0.000000} & 0.203486 \\
\makecell{\backwardsShort} & \textbf{0.798958} & \textbf{0.168883} & 0.028303 & \textbf{0.169140} \\
\bottomrule
\end{tabular}
\end{sc}
\end{small}
\end{center}
\end{table}

\begin{figure}[ht]
\begin{center}
\centerline{\includegraphics[width=\columnwidth]{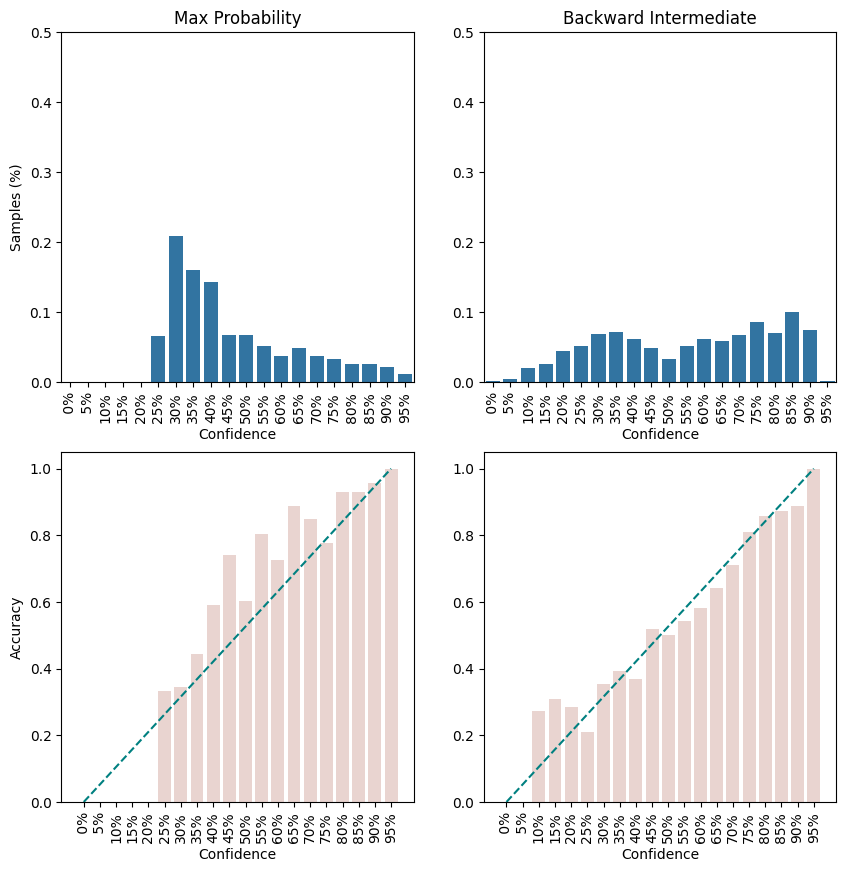}}
\caption{Confidence Histograms and Reliability Diagrams for baseline max probability confidence and internal representations method on the ARC-Challenge dataset.}
\label{arc-c-13b-calib}
\end{center}
\end{figure}

\begin{table}[t]
\caption{Evaluation of Calibration Methods for LLaMA-2 70b model on the ARC-Easy Dataset}
\label{calib-70b-table}
\begin{center}
\begin{small}
\begin{sc}
\begin{tabular}{lcccc}
\midrule
 &  \multicolumn{3}{c}{BoolQ} \\
 \midrule
Method & AUROC$\uparrow$ & Brier $\downarrow$ & ECE$\downarrow$ & smECE$\downarrow$ \\
\midrule
\makecell{\maxPShort} & \textbf{0.703654} & 0.173633 & 0.113462 & 0.191394 \\
\makecell{\entBinShort} & 0.656231 & \textbf{0.163278} &\textbf{ 0.000000} & \textbf{0.164038} \\
\makecell{\auxShort} & 0.475661 & 0.175854 & 0.011057 & 0.176229 \\
\midrule
 &  \multicolumn{3}{c}{ARC(E)} \\
 \midrule
Method & AUROC$\uparrow$ & Brier $\downarrow$ & ECE$\downarrow$ & smECE$\downarrow$ \\
\midrule
\makecell{\maxPShort} & \textbf{0.883857} & 0.082786 & 0.098295 & 0.106471 \\
\makecell{\entBinShort} & 0.870172 & \textbf{0.073242} & \textbf{0.000000} & \textbf{0.074819} \\
\makecell{\auxShort} & 0.575597 & 0.093998 & 0.005770 & 0.093431 \\
\midrule
 &  \multicolumn{3}{c}{ARC(C)} \\
 \midrule
Method & AUROC$\uparrow$ & Brier $\downarrow$ & ECE$\downarrow$ & smECE$\downarrow$ \\
\midrule
\makecell{\maxPShort} & \textbf{0.8324} & 0.1355 & 0.0751 & 0.1500 \\
\makecell{\entBinShort} & 0.8234 & \textbf{0.1319} & \textbf{0.00} & \textbf{0.1332} \\
\makecell{\auxShort} & .5376 & 0.1745 & 0.0273 & 0.1796 \\
\midrule
 &  \multicolumn{3}{c}{MMLU} \\
 \midrule
Method & AUROC$\uparrow$ & Brier $\downarrow$ & ECE$\downarrow$ & smECE$\downarrow$ \\
\midrule
\makecell{\maxPShort} & \textbf{0.799276} & \textbf{0.178071} & 0.027589 & 0.184655 \\
\makecell{\entBinShort} & 0.789645 & 0.179652 & \textbf{0.000000} & \textbf{0.179728} \\
\makecell{\auxShort} & 0.724031 & 0.205717 & 0.022305 & 0.204554 \\
\midrule
 &  \multicolumn{3}{c}{CSQA} \\
 \midrule
Method & AUROC$\uparrow$ & Brier $\downarrow$ & ECE$\downarrow$ & smECE$\downarrow$ \\
\midrule
\makecell{\maxPShort} &  \textbf{0.780704} & 0.197003 & 0.078272 & 0.204572 \\
\makecell{\entBinShort} & 0.760730 & \textbf{0.192085} & \textbf{0.000000} & \textbf{0.191329} \\ 
\makecell{\auxShort} & 0.539460 & 0.242640 & 0.002209 & 0.235307 \\
\bottomrule
\end{tabular}
\end{sc}
\end{small}
\end{center}
\end{table}

\begin{figure}[ht]
\begin{center}
\centerline{\includegraphics[width=\columnwidth\relax]{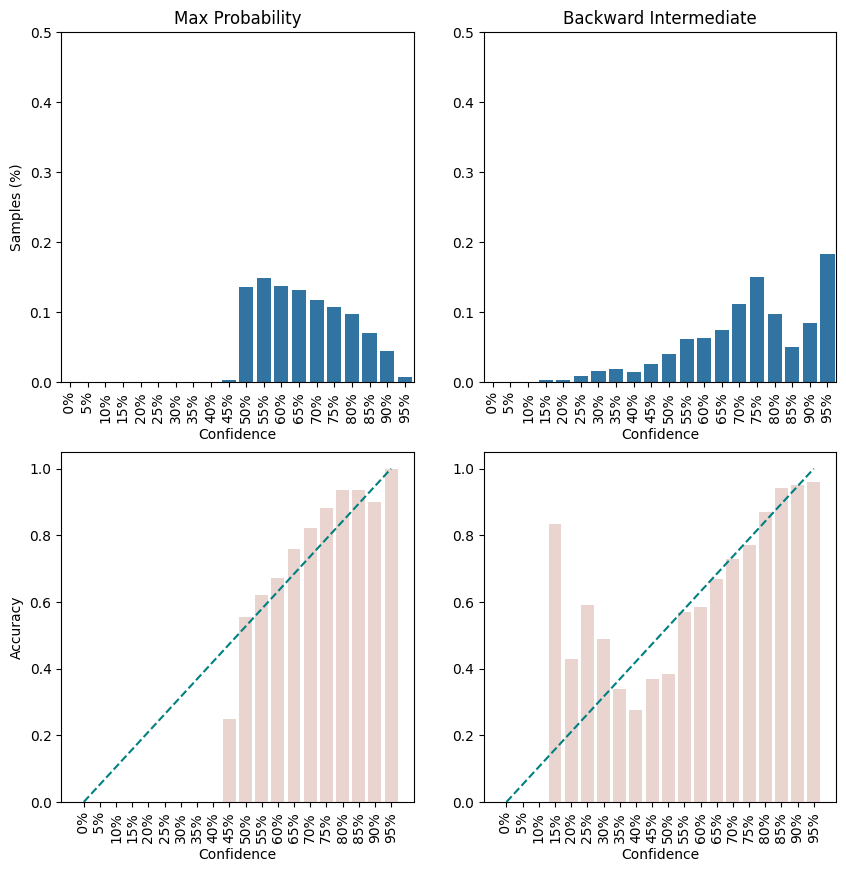}}
\caption{Reliability  diagrams for backwards internal confidence in LLaMA-2 13b on BoolQ.}
\label{calib-boolq}
\end{center}
\end{figure}

\begin{figure}[ht]
\begin{center}
\centerline{\includegraphics[width=\columnwidth\relax]{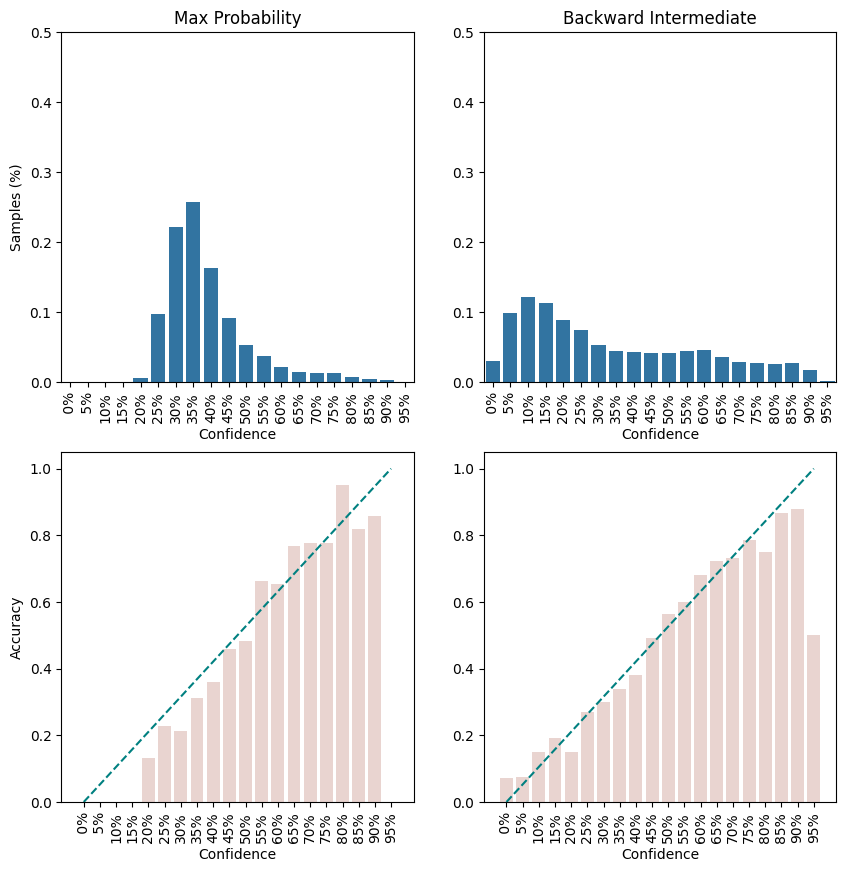}}
\caption{Reliability  diagrams for backwards internal confidence in LLaMA-2 13b on CSQA.}
\label{calib-csqa}
\end{center}
\end{figure}

\section{Oracle Deferral}
\label{app:oracle-deferral}

\begin{figure}[ht]
\begin{center}
\centerline{\includegraphics[width=\columnwidth\relax]{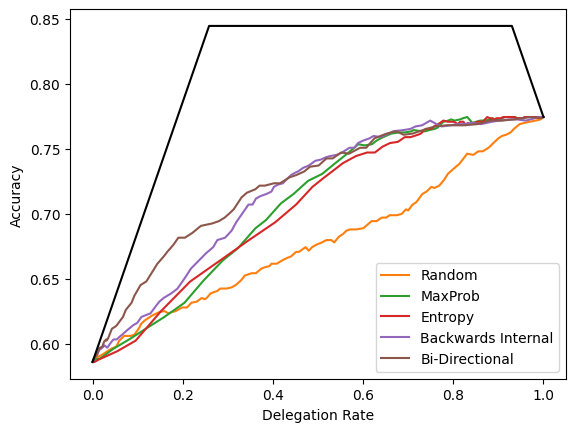}}
\caption{Deferral curves of examined methods on ARC-Challenge contextualized with perfect deferral.}
\label{oracle-curve-arc-c}
\end{center}
\end{figure}

While the cascading approach offers significant performance improvements in many contexts, evaluating the quality of the cascading process is challenging. Total system performance savings, area under the deferral curve, cost-savings at performance parity and other metrics in this domain are sensitive to the individual and relative performance of the models in the cascade on the chosen dataset. Since 100\% task accuracy is often mathematically impossible given the selected models, one method we find helpful in contextualizing deferral performance changes  consideration of the "Oracle Deferral Curve", which we define as the maximum possible task performance at any given deferral rate between two models, without changing the outputs of those models. This perfect oracle initially defers only samples that are correct in the large model and incorrect in the small model until there are no more such samples, then defers samples where the models agree, and finally defers losses until 100\% of samples are deferred. While this quality of deferral is not itself achievable, it represents an upper bound on the deferral process at every rate. \cref{oracle-curve-arc-c} shows the deferral performance of all methods bounded by the perfect deferral curve for the ARC-Challenge dataset.

\section{Auxiliary Forward Model Variant}
\label{app:aux-indiv}

 Table~\ref{table:aux-model-indiv} presents forward calibration performance as well as deferral AUC for two formulations of the auxiliary forward model on MMLU. The first formulation predicts binned accuracy and the alternative formulation predicts individual sample max probability directly produced by the large model. Performance of the two formulations is very similar with slightly higher metrics for the binned approach.

\begin{table*}[t]
\caption{Auxiliary forward model performance for bin accuracy vs individual probability outputs on MMLU with LLaMA-2 70b}
\label{table:aux-model-indiv}
\begin{center}
\begin{small}
\begin{sc}
\begin{tabular}{lccccc}
\toprule
 &  \multicolumn{3}{c}{MMLU} \\
 \midrule
 Method & AUROC$\uparrow$ & Brier Score$\downarrow$ & ECE$\downarrow$ & smECE$\downarrow$ & Deferral AUC$\uparrow$\\
\midrule
\makecell{\binAccShort} & 0.724031 & 0.205717 & 0.022305 & 0.204554 & 0.553650  \\
IndivAcc & 0.720639 & 0.207344 & 0.028835 & 0.208894 & 0.553042 \\
\bottomrule
\end{tabular}
\end{sc}
\end{small}
\end{center}
\end{table*}

\section{Auxiliary Model Judgment Examples}
\label{app:ex}

We present examples of forward confidence estimation in \cref{examples-table}. Examples 1 and 2 belong to the easiest subset of samples. In both cases there is a strong link between the question and the correct answer. The Sun is the primary source of heat in many contexts and thirst is linked to consuming liquids in its definition. Quantitative questions such as Example 3 are considered more difficult, aligning with prior work applying transformer models to simple math questions \cite{Cobbe2021TrainingVT}. Example 4 is considered a sample with very low expected confidence, the correct answer in this case is subtly distinguished from other options that are the subject of ongoing research. 

\begin{table*}[t]
\caption{Example questions and their estimated difficulty according to auxiliary forward confidence model.}
\label{examples-table}
\begin{center}
\begin{small}
\begin{tabular}{p{9cm}ccc}
\toprule
 \multicolumn{4}{c}{\textsc{Easy Question Examples}} \\
 \midrule
 \textsc{Question} &  \textsc{Answer} & \textsc{Predicted Confidence} & \textsc{$M_L$ Correct}\\
\midrule
1. Question: What is the main source of heat for Earth's surface? 
Choices: (A): fire (B): lightning (C): the Sun (D): the ocean & C & High & Yes \\
\midrule
2. Question: When people exercise, they often feel thirsty and begin to sweat. It is important for people to feel thirsty when exercising because it makes them realize that they should 
Choices: (A): take a break (B): consume liquids (C): slow their breathing (D): stop to eat something & B & High & Yes \\
 \midrule
  \multicolumn{4}{c}{\textsc{Hard Question Examples}} \\
 \midrule
 3. Question: A potassium (K) atom has 20 neutrons, 19 protons and 19 electrons. What is the atomic mass of potassium? 
 Choices: (A): 19 (B): 20 (C): 38 (D): 39 & D & Low & No \\
 \midrule
 4. Question: Which situation is an example of an inherited trait? 
 Choices: (A): lions preying on zebras (B): monkeys using twigs to get food (C): birds following migratory patterns (D): bears opening coolers at campsites & A & Low & No \\
\bottomrule
\end{tabular}
\end{small}
\end{center}
\end{table*}

\section{Calibration for Llama 7b}
\label{app:calibration-llama-7b}

\begin{figure}[ht]
\begin{center}
\centerline{\includegraphics[width=\columnwidth\relax]{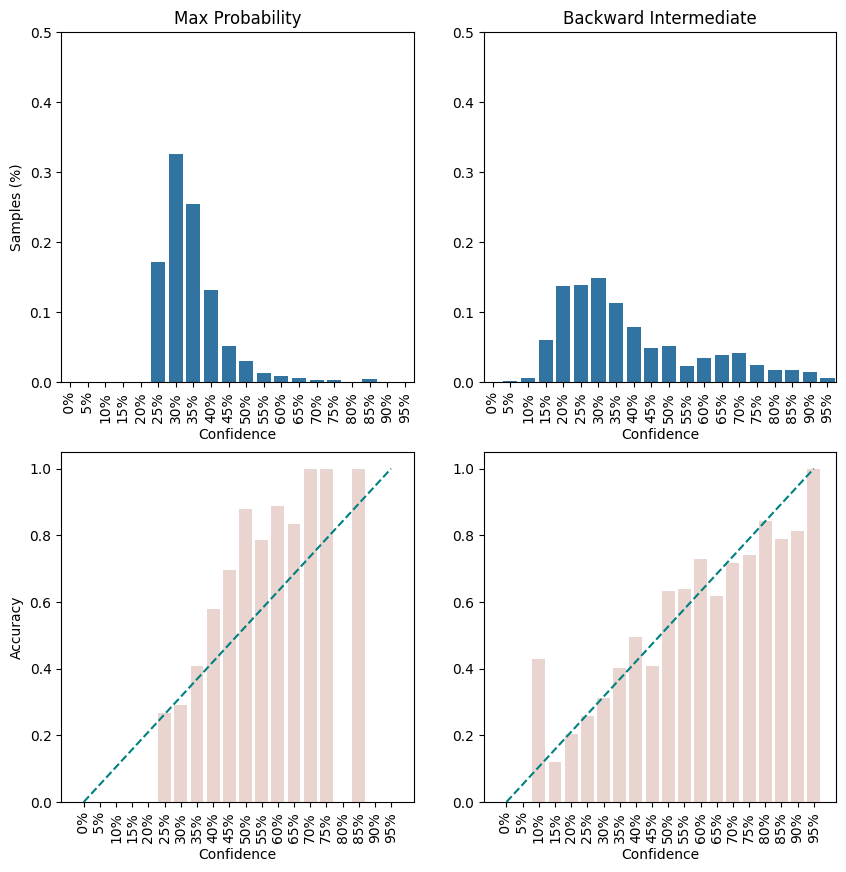}}
\caption{Calibration performance for LLaMA-2 7b on ARC-Challenge.}
\label{7b-calibration}
\end{center}
\end{figure}

\begin{figure}[ht]
\begin{center}
\centerline{\includegraphics[width=\columnwidth\relax]{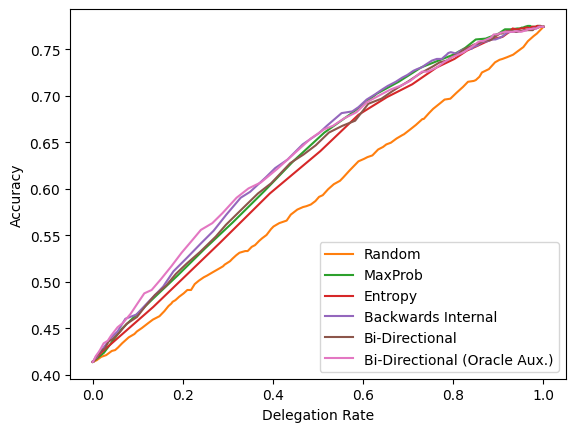}}
\caption{Deferral curves for cascading with LLaMA-2 7b as $M_S$ on ARC-Challenge.}
\label{7b-deferral}
\end{center}
\end{figure}

In the case of LLaMA-2 7b for ARC-Challenge and other difficult tasks, confidence reflects generally near-chance accuracy for a majority of samples, as indicated in \cref{7b-calibration}. This near chance estimate, while appropriate given the observed performance, limits the viability of intelligent deferral and weakens cascade performance for both baselines and our methods as shown in \cref{7b-deferral}.

\section{Deferred Prompt Length}
\label{app:prompt-length}

While average prompt lengths are used for estimating relative cost reductions of methods, a more detailed analysis of which samples are deferred under different methods revealed that the baseline method delegates longer samples to the larger model in most cases compared to our method. This means that relative cost reductions are in fact larger than estimated, since computational cost is a function of prompt length and deferral of larger prompts is more costly. Table ~\ref{prompt-len-table} presents the mean prompt length (in tokens) for samples sent to the larger model using the \bidirShort vs \maxPShort at a .2 deferral rate. The deferral based solely on confidence of the small model may prioritize deferral of long, complex and difficult samples to the large model leading to increased cost per deferred sample. While deferring difficult samples is generally a good idea, a more nuanced approach to deferral can prioritize deferral of samples that are likely to be correct in the large model and avoid wasteful deferral of complex samples that both models will predict incorrectly.

\begin{table*}[t]
\caption{Mean prompt length at .2 deferral rate for MaxProb and Bi-directional deferral and average over all datasets.}
\label{prompt-len-table}
\begin{center}
\begin{small}
\begin{sc}
\begin{tabular}{lrrrrr|r}
\toprule
 & BoolQ & ARC(E) & ARC(C) & MMLU & CSQA & Average \\
Method &  &  &  &  &  \\
\midrule
\makecell{\maxPShort} & 648.88 & 275.4 & 288.38 & 544.177 & 169.676 & 385.30\\
\makecell{\bidirShort} & 629.74 & 270.88 & 305.51 &  502.74 & 170.66 & 375.90 \\
\bottomrule
\end{tabular}
\end{sc}
\end{small}
\end{center}
\end{table*}

\section{Bootstrap Confidence Intervals}
\label{app:bootstrap-confidence}

To measure the consistency of the improvements over baselines we conduct a non-parametric statistical test of significance using bootstrap resampling and report improvements over MaxProb at all examined operating points. Test sets were resampled with replacement 1000 times and performance difference at the 95\% confidence interval (CI) is reported. CIs do not include zero for any operating point or dataset for the BiDir method.

\begin{table*}[t]
\caption{Bi-Dir AUC improvement over baseline (MaxProb) 95\% bootstrap CI .}
\label{bidir-bootstrap-table}
\begin{center}
\begin{small}
\begin{sc}
\begin{tabular}{llrrrl}
\toprule
Dataset & Operating Point & Mean & CI Low & CI High &	Significant \\
\midrule
 &	AUC (.2) &	.001324 &	.000071 &	.002595 &	YES \\
BoolQ &	AUC (.4) &	.003924 &	.001668 &	.006296 &	YES \\
 &	AUC (full) &	.008550 &	.005581 &	.011764 &	YES \\
 \midrule
 &	AUC (.2) &	.002191 &	.000921 &	.003475 &	YES \\
ARC-Easy &	AUC (.4) &	.005556 &	.003433 &	.007666 &	YES \\
 &	AUC (full) &	.005570 &	.002876 &	.008313 &	YES \\
  \midrule
 & AUC (.2) &	.005068 &	.002680 &	.007653 &	YES \\
ARC-Challenge &	AUC (.4) &	.012138 &	.006412 &	.017708 &	YES \\
 &	AUC (full) &	.014041 &	.006127 &	.021983 &	YES \\
  \midrule
 &	AUC (.2) &	.002370 &	.001794 &	.002973 &	YES \\
MMLU &	AUC (.4) &	.007260 &	.005666 &	.008851 &	YES \\
 &	AUC (full) &	.013152 &	.010458 &	.015871 &	YES \\
  \midrule
 &	AUC (.2) &	.002159 &	.000841 &	.003349 &	YES \\
CSQA &	AUC (.4) &	.005165 &	.002423 &	.007724 &	YES \\
 &	AUC (full) &	.014128 &	.009707 &	.018671 &	 YES \\
\bottomrule
\end{tabular}
\end{sc}
\end{small}
\end{center}
\end{table*}

\begin{table*}[t]
\caption{BackInt AUC improvement over baseline (MaxProb) 95\% bootstrap CI .}
\label{backint-bootstrap-table}
\begin{center}
\begin{small}
\begin{sc}
\begin{tabular}{llrrrl}
\toprule
Dataset & Operating Point & Mean & CI Low & CI High &	Significant \\
\midrule
 &	AUC (.2) &	.002365 &	.001240 &	.003479 &	YES \\
BoolQ &	AUC (.4) &	.005974 &	.003948 &	.007874 &	YES \\
 &	AUC (full) &	.013061 &	.009767 &	.016529 &	YES \\
 \midrule
 &	AUC (.2) &	.001806 &	.000884 &	.002811 &	YES \\
ARC-Easy &	AUC (.4) &	.004700 &	.002960 &	.006467 &	YES \\
 &	AUC (full) &	.003923 &	.001181 &	.006552 &	YES \\
  \midrule
 & AUC (.2) &	.001825 &	.000406 &	.003144 &	YES \\
ARC-Challenge &	AUC (.4) &	.005729 &	.002393 &	.008867 &	YES \\
 &	AUC (full) &	.007969 &	.002361 &	.013233 &	YES \\
  \midrule
 &	AUC (.2) &	.000003 &	-.000453 &	.000514 &	NO \\
MMLU &	AUC (.4) &	.001280 &	.000285 &	.002226 &	YES \\
 &	AUC (full) &	.003362 &	.001816 &	.004953 &	YES \\
  \midrule
 &	AUC (.2) &	.002028 &	.000766 &	.003241 &	YES \\
CSQA &	AUC (.4) &	.003900 &	.000993 &	.006784 &	YES \\
 &	AUC (full) &	.016225 &	.010289 &	.021932 &	 YES \\
\bottomrule
\end{tabular}
\end{sc}
\end{small}
\end{center}
\end{table*}

\section{Limitations}
\label{sec:limitations}
In this work we evaluate LLM cascades using only models from the LLaMA-2 series due to their recency, transparency and accessibility. While this choice reduces confounds and facilitates reproducibility, there are some limitations as a result. Larger models may better evaluate the value of cascades in very difficult tasks, the performance of our small model on difficult tasks shows a significant amount of ``guessing'', while this is expected for the smaller model it introduces an element of chance that is not conducive to consistent results. There is also some evidence that diversity of component models in a cascade may improve its performance which we are unable to explore with this model set \cite{Lebovitz2023EfficientIW}. 

In this work we use zero-shot prompting, avoiding multi-shot and multi-output prompting strategies where the computational cost is not a simple function of the input and output length. This allows a simple comparison of costs between models of different sizes but lowers the individual model performance compared to claimed values particularly in smaller models. In some cases using alternative prompting strategies in a cascade may yield better efficiency gains than our cascades with consistent, simple prompting, however this introduces additional degrees of variability and is not the focus of this work.

\end{document}